\documentclass[preprint,5p,times,twocolumn]{elsarticle}

\usepackage{subfigure}
\usepackage{amssymb}
\usepackage{amsmath}
\usepackage{algorithm}
\usepackage{algorithmic}

\journal{Knowledge-Based Systems}
\usepackage{lineno}
\begin{document}
\begin{frontmatter}



\title{FedPURIN: Programmed Update and Reduced INformation for Sparse Personalized Federated Learning}


\author[add1]{Lunchen Xie}
\author[add1]{Zehua He}
\author[add1,add2]{Qingjiang Shi\corref{cor1}}
\cortext[cor1]{Corresponding author}\ead{shiqj@tongji.edu.cn}

\address[add1]{School of Computer Science and Technology, Tongji University, No.4800 Cao'an Hwy, Jiading District, Shanghai, 201804, China}
\address[add2]{Shenzhen Research Institute of Big Data, 2001 Longxiang Avenue, Longgang District, Shenzhen, 518172, China}

\begin{abstract}
Personalized Federated Learning (PFL) has emerged as a critical research frontier addressing data heterogeneity issue across distributed clients. Novel model architectures and collaboration mechanisms are engineered to accommodate statistical disparities while producing client-specific models. Parameter decoupling represents a promising paradigm for maintaining model performance in PFL frameworks. However, the communication efficiency of many existing methods remains suboptimal, sustaining substantial communication burdens that impede practical deployment. To bridge this gap, we propose Federated Learning with Programmed Update and Reduced INformation (FedPURIN), a novel framework that strategically identifies critical parameters for transmission through an integer programming formulation. This mathematically grounded strategy is seamlessly integrated into a sparse aggregation scheme, achieving a significant communication reduction while preserving the efficacy. Comprehensive evaluations on standard image classification benchmarks under varied non-IID conditions demonstrate competitive performance relative to state-of-the-art methods, coupled with quantifiable communication reduction through sparse aggregation. The framework establishes a new paradigm for communication-efficient PFL, particularly advantageous for edge intelligence systems operating with heterogeneous data sources.
\end{abstract}




\end{frontmatter}



\newcommand{\btheta}{\boldsymbol{\theta}}
\newcommand{\bm}{\mathbf{m}}
\newcommand{\bg}{\mathbf{g}}
\newcommand{\mD}{\mathcal{D}}

\section{Introduction}
Federated learning (FL), as a powerful distributed machine learning scheme, has been well studied to handle the growing trend towards harnessing abundant data on ubiquitous edge devices~\cite{mcmahan2017communication}. This framework has been successfully applied in various domains, including computer vision~\cite{xu2023personalized, chen2022bridging}, healthcare~\cite{rieke2020future, srinivasu2024enhancing}, finance~\cite{long2020federated, myalil2021robust}, and ubiquitous IoT applications~\cite{duan2019jointrec, yin2020fdc, chai2020hierarchical}. A dominant paradigm, exemplified by the seminal FedAvg algorithm, seeks a single consensus model across all participants through the iterative exchange and aggregation of intermediate model updates. This process allows clients to share learned knowledge in a distributed manner without leaking raw data, enabling them to collectively improve model performance and accelerating practical deployment. However, multi-source data distributions create significant collaboration barriers, intensifying the data-silo dilemma~\cite{yang2019federated}. This inherent data heterogeneity across clients critically undermines the efficacy of conventional FL. Specifically, the statistical disparity causes the local objective functions to diverge from the global optimum, leading to the well-known problem of client drift. 
This situation creates a fundamental conflict between the uniformity of a global model and the personalization required for domain-specific tasks.

To address this problem, personalized federated learning (PFL) is gaining attention from both academia and industry~\cite{fallah2020personalized, li2021ditto, t2020personalized}. Unlike conventional FL that seeks a single global model, PFL aims to learn a set of personalized models tailored to individual clients' data distributions. The special design of collaboration strategies allows PFL algorithms to handle the distribution isomerism of personal information effectively. These strategies can be broadly categorized into several paradigms. For example, FedPer method~\cite{arivazhagan2019} shields the critical classification layer from global aggregation, achieving an apparent performance increase. FedRep~\cite{collins2021exploiting} also discusses the importance of building a shared data representation but maintaining unique local model segments. This parameter-decoupling architecture effectively mitigates the negative interference from non-IID data. Different from these parameter decoupling strategies, seeking help from knowledge distillation is applicable as well. The representative~\cite{jin2022personalized} obtains the mixed model through both local model update and global knowledge alignment. Additionally, a more radical approach to personalized collaboration involves splitting the model and transferring intermediate activations for cross-client knowledge distillation~\cite{he2020group}, which notably increases the communication burden compared to parameter-only exchanges.

While existing PFL approaches advance personalization, they introduce persistent bottlenecks in communication efficiency and collaborative optimization. This necessitates a fundamental redesign of parameter selection and decoupling mechanisms, which is a rising research frontier where traditional deep learning pruning methods offer limited solutions when adapted to federated settings. This is primarily because conventional pruning techniques are designed for centralized data, which fail to account for the statistical heterogeneity inherent in federated environments. Investigation of the contribution or sensitivity of the parameters~\cite{molchanov2019importance, zhang2022pats} provides a way to assess the role of different parameters. Although FedCAC~\cite{wu2023bold} pioneers parameter decoupling for PFL training, it relies on heuristics rather than a rigorous optimization framework for parameter assessment and still requires full-model updates. This leads to substantial communication overhead, impeding practical deployment on resource-constrained edge devices. These limitations highlight the need for a new optimization-theoretic framework that not only provides principled mathematical grounds for parameter decoupling, but also achieves substantial communication reduction with acceptable computational cost.

In this paper, we propose Federated Learning with Programmed Update and Reduced INformation (FedPURIN\footnote{The source code is available at: https://github.com/HikariX/FedPURIN}), a novel PFL framework that simultaneously enhances edge model personalization and reduces communication overhead.
FedPURIN introduces a theoretically grounded parameter decoupling mechanism based on integer programming, which dynamically evaluates parameter importance and generates binary masks to identify critical parameters under non-IID data distributions. This formulation enables the server to aggregate models using only the uploaded critical parameters, effectively suppressing communication overhead by 46\% to 73\% across various scenarios while maintaining competitive model performance. Throughout this process, each client's personalized model is strategically composed of both specially collaborated critical parameters and trivially aggregated non-critical parameters, creating an adaptive architecture that dynamically adjusts to varying data distributions and client requirements. The main contributions of this study are summarized as follows:

\begin{itemize}
\item We establish a novel parameter decoupling perspective via integer programming to evaluate the parameter importance in PFL problem, thereby forming the theoretical foundation of FedPURIN.
\item Building upon this metric, we develop a communication-efficient protocol that transmits exclusively critical parameters, reducing communication overhead by 46\% to 73\% across scenarios.
\item Extensive experiments across diverse datasets and non-IID settings conclusively validate the effectiveness and efficiency of FedPURIN, with in-depth component analyses providing strong evidence for its efficacy.
\end{itemize}

\begin{figure*}[!t]
    \centering
    \includegraphics[width=1\linewidth]{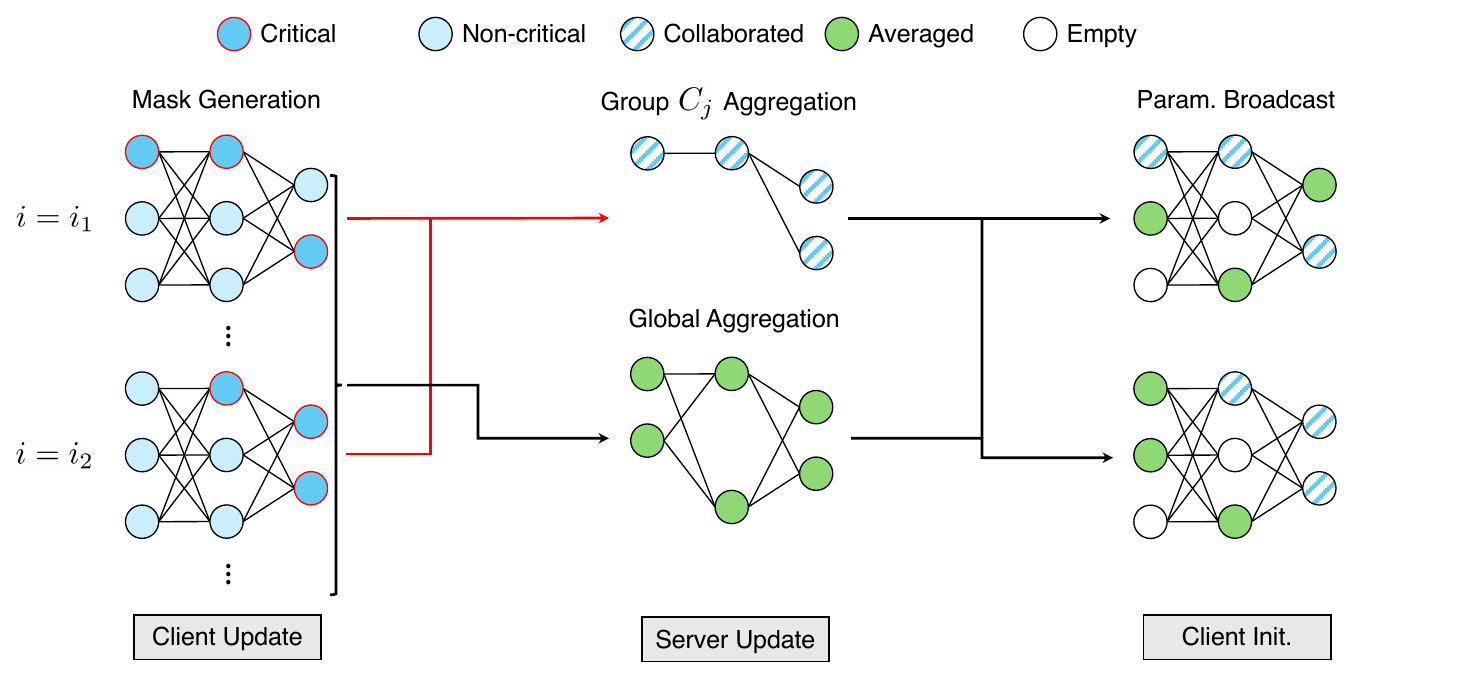}
    \caption{The workflow of FedPURIN algorithm in communication round $t$. There are mainly four steps: (1) Clients conduct local training to determine critical parameter sets, then upload them; (2) Clients with high similarity on the critical parameters masks are grouped to generate the collaborated weights; (3) Server collects critical parameters from every client to form the trivial global model through FedAvg scheme; (4) Server send back the collaborated/trivial weights for critical/non-critical parameters.}
    \label{fig:pipeline}
\end{figure*}

\section{Related Work}
\subsection{Personalized Federated Learning}
To address the fundamental challenge of data heterogeneity in federated learning, PFL has emerged as a pivotal research direction. Existing approaches can be broadly categorized into several paradigms based on their underlying strategies. Alternating approaches~\cite{chen2022bridging, zhang2023gpfl} decouple the model into personalized and shared components, then cyclically freeze one part while training the other. Shared-representation learning~\cite{arivazhagan2019, collins2021exploiting, li2021fedbn, pillutla2022federated}, identifies specific modules (like classifiers) that are kept local, and fulfills the collaboration by excluding them from aggregation. Split Learning~\cite{vepakomma2018split, thapa2022splitfed} operates on a different principle, where clients transmit intermediate representations (e.g., activation maps) to the server. While these are not raw data, the communication cost of transferring these high-dimensional activations can be substantial, often surpassing that of parameter exchanges. In contrast, knowledge transfer methods aim to make clients and the server learn from each other rather than merely share the information. Techniques in this category achieve the goal by aligning local model parameters with the server and previous local ones ~\cite{jin2022personalized}, or by leveraging intermediate outputs for mutual distillation~\cite{he2020group, wu2023fedict, atapour2024fedd2s}. This paradigm focuses on achieving consensus through the learned knowledge rather than the fusion of model components or intermediate features. Collectively, these methods advance personalization but often incur significant communication costs, either from transmitting parameters, intermediate activations, or synthetic data. This persistent overhead underscores the need for more communication-efficient PFL frameworks.

\subsection{Parameter Selection Criteria}
Identifying and selecting influential parameters is crucial for achieving efficient collaboration in PFL. The key challenge lies in establishing effective criteria to determine which parameters are most valuable for collaborative training under various data distributions. In centralized deep learning, parameter selection techniques such as pruning serve as regularization methods to prevent overfitting and reduce model size~\cite{molchanov2019importance, zhang2022pats}. However, the direct application of these centralized parameter selection criteria to federated settings faces significant challenges, as they typically assume homogeneous data distributions and fail to account for the statistical heterogeneity across clients. When adapted to FL, this approach transforms into a strategic parameter selection mechanism, selectively identifying and transmitting the most influential parameters for collaboration. This adaptation therefore requires carefully designed strategies and metrics to balance local and global learning, a challenge that is not yet fully resolved.

The core of parameter selection lies in the metric used to evaluate the importance of the parameters. Early attempts often adopted straightforward heuristics by randomly selecting a fixed sub-network for each client to train and collaborate~\cite{munir2021fedprune}, lacking a principled criterion for parameter selection. Subsequent work such as FedSelect~\cite{tamirisa2024fedselect} relies on relatively simple heuristics like selecting parameters based solely on the magnitude of their local updates. Building on this, foundational work on parameter selection like FedPrune~\cite{liu2021fedprune} introduced the concept of perturbation to adaptively examine the sensitivity of parameters across network layers. This line of research was further advanced by FedCAC~\cite{wu2023bold}, which contributes to a more structured collaboration scheme by grouping and aggregating critical parameters separately, leading to improved performance. Although these methods have significantly advanced the field, their strategies for parameter selection and evaluation open up opportunities for further exploration under a unified optimization-theoretic framework. Establishing such a framework could provide a more generalizable foundation for dynamically adapting to the full spectrum of client data heterogeneity, which represents a promising direction for intelligent parameter selection toward sparse collaboration in PFL.

\section{Method}
This section details our proposed FedPURIN with the elaboration of integer-programming enpowered parameter decoupling method and sparse model aggregation strategy. The overall framework is illustrated in Fig.~\ref{fig:pipeline}
\subsection{Problem Formulation}
Conventional FL algorithms (e.g., FedAvg~\cite{mcmahan2017communication}) aim to learn a single global consensus model that performs reasonably well across all participating clients. However, due to the inherent statistical heterogeneity (non-IID data distribution), a single global model often fails to optimally fit the specific data characteristics of each individual client. This mismatch becomes particularly pronounced when local data distributions diverge significantly, as the global model tends to converge to a compromised solution that does not excel in any specific local context. To address this limitation, PFL methods seek to learn a set of personalized models, each tailored to the unique data distribution and requirements of its respective client. Correspondingly, the objective of the PFL problem is typically formulated as follows:
\begin{equation}\label{eq1}
    \min_{\btheta_1, \btheta_2,...,  \btheta_N} \frac{1}{N}\sum_{i=1}^{N}L_i(\btheta_i, \mD_i),
\end{equation}
where $N$ denotes the number of clients, $\mD_i$ and $\btheta_i$ mark the dataset and model of $i$-th client respectively. In contrast to conventional FL which seeks a universal model, each client $i$ builds its model upon its personal dataset $\mD_i$. This formulation presents a fundamental challenge: how to enable effective collaboration among clients while preserving their individual model specificity. In each communication round, the server typically selects a subset of clients. Each selected client $i$ then performs local training and computes updates to minimize its local objective function $L_i(\cdot)$. During this process, clients still cooperate on their model parameters $\btheta_i$, making the optimization problem particularly challenging due to the dual requirements of personalization and collaboration. Therefore, the key to this challenge lies in developing a principled method to identify and collaborate on the right set of parameters which benefit from aggregation without compromising local specificity.

\subsection{Parameter Decoupling -- A QIP Perspective}\label{sec:3.2}
Each client's model parameters encode knowledge derived from its local data. For the PFL task, this knowledge becomes specialized to client-specific patterns within distinct parameter subsets. Existing approaches to this problem often rely on dedicated mechanisms or heuristic criteria to isolate and protect personalized parameter components. While these methods advance personalization, their identification of critical parameters frequently lacks a rigorous optimization foundation, and more critically, many still necessitate a full-model upload, sustaining a substantial communication burden.

To address these dual limitations of theoretical rigor and communication efficiency, we propose FedPURIN, a novel method that applies a binary masking mechanism to preserve important parameters while zeroizing non-critical ones. During each communication round, clients upload only the sparse set of non-masked parameters (termed critical parameters). The server then aggregates these values via careful averaging and redistributes the updated critical parameters. This intentional decoupling induces dynamic perturbation to the model. To ensure robustness, we first elaborate the perturbation effects and decoupling strategy here, followed by a detailed description of the sparse aggregation scheme in the next subsection.

To rigorously formalize this mechanism, a binary mask $\bm_i$ is introduced for each client $i$, where the elements corresponding to the specified parameters are 0, otherwise 1. The loss function can be extended from $L_i(\btheta_i, \mD_i)$ to $L_i(\btheta_i\odot\bm_i, \mD_i)$, thus we can derive the loss function of the $i$-th client in the communication round $t+1$ with the Taylor approximation as follows:
\begin{equation}\label{eq2}
\begin{aligned}
    L(\btheta\odot\bm)\approx
    &L(\btheta\odot\bm^{(t)})+\langle\bg,(\btheta\odot\bm)-(\btheta\odot\bm^{(t)})\rangle\\
    &+\frac{1}{2}(\btheta\odot\bm-\btheta\odot\bm^{(t)})^TH(\btheta\odot\bm-\btheta\odot\bm^{(t)}),
\end{aligned}
\end{equation}
where $\bg=\nabla L$ is the gradient and $H$ is the Hessian. For simplicity, the subscripts denoting the $i$-th client are omitted here.

In FedPURIN, clients decouple full models via unstructured sparsification before upload. The server aggregates these parameters and builds the new model for the next local training. Since local models retain full parameters before decoupling, the previous mask can be viewed as $\bm^{(t)} = \mathbf{1}$. The core task is thus simplified to deactivating a subset of parameters by setting the corresponding mask entries to zero. This renders the first term of Eq.~\eqref{eq2} constant, and hense omissible. Therefore, the minimization of the local loss function is transformed into a Quadratic Integer Programming (QIP) problem:
\begin{equation}\label{eq3}
\begin{aligned}
    &\min\big|\sum_{j=1}^d [\bg]_j[\btheta]_j([\bm]_j-[\bm^{(t)}]_j)\\
    &+\frac{1}{2}\sum_{j=1}^d\sum_{k=1}^d[H]_{j,k}[\btheta]_j[\btheta]_k([\bm]_j-[\bm^{(t)}]_j)([\bm]_k-[\bm^{(t)}]_k)\big|,
\end{aligned}
\end{equation}
where $d$ is the total number of parameters, and $[\cdot]j$ denotes the $j$-th element of a variable. The direct computation of the Hessian matrix is computationally expensive. To address this, two common approximations are employed. First, the Hessian is simplified via the Becker–LeCun approximation~\cite{becker1988improving}, which discards off-diagonal terms $[H]_{j,k} (j\neq k)$. Second, the Hessian is further approximated by the Fisher information matrix (FIM)~\cite{martens2020new}. For a mean-reduced loss $L(\btheta)=\frac{1}{B}\sum_bl(f(x_b,\btheta),y_b)$, the $j$-th diagonal element of the Hessian is
\begin{equation}\label{eq4}
[H(\btheta)]_{j,j} = \frac{1}{B}\sum_{b=1}^B\frac{\partial^2 l(f(x_b,\btheta),y_b)}{\partial [\btheta]_j^2},
\end{equation}
where $x_b$ and $y_b$ denote the input sample and its label, and $B$ is the batch size. In practice, we use the empirical FIM, where the expectation is taken over training samples rather than the model distribution~\cite{kunstner2019limitations}. The corresponding diagonal element of the empirical FIM over the same batch is
\begin{equation}\label{eq5}
[F]_{j,j} = \mathbb{E}[(\frac{\partial L}{\partial [\btheta]_j})^2]=\frac{1}{B}\sum_{b=1}^B(\frac{\partial l(f(x_b,\btheta),y_b)}{\partial [\btheta]_j})^2.
\end{equation}
This substitution is theoretically justified when the model’s predictive distribution closely matches the empirical data distribution, which often holds near convergence.

In practice, computing per-sample gradients for the above quantity is computationally demanding. Hence, a lighter approximation is adopted as $[\tilde{F}]_{j,j} =(\frac{1}{B}\sum_{b}\frac{\partial l}{\partial [\btheta]_j})^2$, which corresponds to the square of the batch-averaged gradient $[\bg]_j$ and can be directly obtained from PyTorch implementations. This approximation assumes that the within-batch variance of per-sample gradients is negligible--an assumption that is empirically valid in large-batch training and commonly employed in adaptive optimizers such as RMSProp~\cite{tieleman2012lecture} and Adam~\cite{kingma2014adam}. Although this formulation tends to slightly underestimate the true curvature magnitude, it provides a stable and computationally efficient diagonal curvature proxy that has been widely validated in practice. Combining these approximations leads to the following simplified problem:
\begin{equation}\label{eq6}
\begin{aligned}
\min&\big|\sum_{j=1}^d[\bg]_j[\btheta]_j([\bm]_j-[\bm^{(t)}]_j)\\
  &+\frac{1}{2}\sum_{j=1}^d[\bg]_j^2[\btheta]_j^2([\bm]_j-[\bm^{(t)}]_j)^2\big|.
\end{aligned}
\end{equation}

For the $j$-th element, the perturbation varies in two situations:
\begin{itemize}
    \item Mask unchanged ($[\bm]_j=[\bm^{(t)}]_j$): the perturbation is 0.
    \item Mask flipped ($[\bm]_j=1-[\bm^{(t)}]_j$): the perturbation is 
    \begin{equation}\label{eq7}
    |[\bg]_j[\btheta]_j(1-2[\bm^{(t)}]_j)+\frac{1}{2}[\bg]_j^2[\btheta]_j^2(1-2[\bm^{(t)}]_j)^2|.
    \end{equation}
\end{itemize}
Recall that the model retains full parameter before global aggregation, so all entries of the previous mask are ones. Thus, the perturbation reduces to $|-[\bg]_j[\btheta]_j+\frac{1}{2}[\bg]_j^2[\btheta]_j^2|$. We also notice that this expression is intrinsically equivalent to the sensitivity measure $|[\nabla_{\btheta}L]_j[\btheta]_j|$ as proposed in the previous method \cite{wu2023bold}, which omits the second-order term. In implementation, $\bg$ can be replaced by the variation of the model parameters $\Delta\btheta$ for simplicity.

With the crafted perturbation value as the metric, FedPURIN needs to build binary masks for different clients. After the end epoch of local training, each client will examine the values layer by layer.  We use $\tau\in[0,1]$ as the hyperparameter to control the picking proportion of parameters for global aggregation. The mask $\bm$ recording the position of critical parameters of one client includes:
\begin{equation}\label{eq8}
    [\bm]_j=
    \begin{cases} 
    1, & \text{if } [\bm]_j \text{ in the top-}\tau \text{ largest of } \bm\\
    0, & \text{otherwise }
\end{cases}
\end{equation}

\subsection{Communication-Efficient Parameter Collaboration}
\label{sec:3.3}
Once the server receives various critical parameters, a proper collaboration strategy is required to extract common knowledge. Inspired by FedCAC~\cite{wu2023bold}, we also fulfill the collaboration of critical parameters between clients $i$ and $j$ by calculating the overlap ratio $O_{i,j}^{(t)}=1-\frac{||\bm_i^{(t)}-\bm_j^{(t)}||_1}{2n}$, where $||\cdot||_1$ is the $L_1$ norm and $n$ denotes the total number of critical parameters of a client. After that, a threshold $\mathcal{T}^{(t)}=O_{avg}^{(t)}+\frac{t}{\beta}\times(O_{max}^{(t)}-O_{avg}^{(t)})$ is calculated, where $O_{avg}^{(t)}=\frac{1}{N(N-1)}\sum_{i\neq j}O_{i,j}^{(t)}$ and $O_{max}^{(t)}=\max_{i\neq j}\{O_{i,j}^{(t)}\}$. For client $i$, the collaboration set is defined by $C_i^{(t)}=\{j|O_{i,j}^{(t)}\geq\mathcal{T}^{(t)},j\neq i\}$, and the collaborated model is
\begin{equation}\label{eq9}
    \boldsymbol{\delta}_i^{(t)}=\frac{1}{|C_i^{(t)}+1|}\sum_{j\in C_i^{(t)}\cup{i}}\btheta_j^{(t)}.
\end{equation}
The value of $\mathcal{T}^{(t)}$ rises as the training proceeds, resulting in the independent critical set for each client when $t>\beta$.

Although FedCAC elaborates the applicable collaboration scheme, it still demands the participation of non-critical parameters to build the global model $\bar{\btheta}^{(t)}=\frac{1}{N}\sum_{i=1}^N\btheta_i^{(t)}$ for the incoming updates. This design inherently limits its communication efficiency, as every parameter must be transmitted in each round, regardless of its assessed importance. In addition, the final model ready for each client is the combination of $\boldsymbol{\delta}_i^{(t)}$ and $\bar{\btheta}^{(t)}$. It is similar to traditional FL methods requiring the full-model update, which poses high communication overhead between the client and the server. Consequently, this strategy causes excessive communication cost and raises our curiosity: Are the critical parameters enough for collaboration? More precisely, can a sparser representation, involving only the critical set, suffice to maintain collaborative learning performance?

Our investigation confirms this hypothesis. Indeed, the calculation of perturbation implies the capability of sparsified models for collaboration, as only critical parameters maintain values during transmission. Moreover, the dynamic nature of binary masks after each collaboration ensures that all model elements retain the opportunity to be selected as critical ones. This dynamic selection prevents any parameter from being permanently silenced, thereby maintaining the model's expressive capacity over time. The temporary exclusion of non-critical parameters imposes no detrimental effect on the final performance. In fact, it can be viewed as a form of regularization, reducing interference from client-specific noise. Building upon these synergistic insights, the FedPURIN framework achieves a communication-efficient aggregation through a simple yet powerful mechanism on the global model:
\begin{equation}\label{eq10}
    \bar{\btheta}^{(t)}=\frac{1}{N}\sum_{i=1}^N\btheta_i^{(t)}\odot\bm_i^{(t)}.
\end{equation}
Before each aggregation, the binary masks is applied to each local model, forcing the non-critical ones to 0 before upload. When the server gathers different model parameters, it updates both the critical parts and the rest. During training, the zeroized model $\btheta_i^{(t)}\odot\bm_i^{(t)}$ is more communication-efficient. By integrating Eq.~\eqref{eq9} and Eq.~\eqref{eq10}, the combined personalized model of client $i$ after aggregation is
\begin{equation}\label{eq11}
    \btheta_i^{(t+1)}=\boldsymbol{\delta}_i^{(t)}\odot \bm_i^{(t)}+\bar{\btheta}^{(t)}\odot\lnot\bm_i^{(t)},
\end{equation}
where $\lnot\bm_i^{(t)}$ is the reverse mask to indicate non-critical parameters. The combined model serves as the starting point for subsequent local trainings. Through this strategy, FedPURIN achieves dual objectives: enabling communication-efficient collaboration across all parameters via sparse aggregation while rigorously preserving client-specific knowledge encoded in local parameters. The complete training procedure is formalized in Algorithm~\ref{alg:purin}.

\begin{algorithm}[tb]
\caption{FedPurin}
\label{alg:purin}
\textbf{Input}: Each client $i$'s model $\btheta_i^{(0)}$; Number of clients $N$; Total communication round $T$; Local epoch number $E$; Hyperparameters $\tau$, $\beta$.\\
\textbf{Output}: Model $\btheta_i^{(T)}$ for client $i$.\\
\vspace{-5mm}
\begin{algorithmic}[1] 
\FOR{$t=1$ to $T$}
\STATE \textbf{Client-side}:
\FOR{$i=1$ to $N$ in parallel}
\STATE Update $\btheta_i^{(t)}$ for $E$ local epochs.
\STATE Evaluate the perturbation $|-[\bg_i]_j[\btheta_i^{(t)}]_j+\frac{1}{2}[\bg_i]_j^2[\btheta_i^{(t)}]_j^2|$ for each $j$-th element.
\STATE Obtain critical parameter mask $\bm_i^{(t)}$.
\STATE Send $\btheta_i^{(t)}\odot\bm_i^{(t)}$ and $\bm_i^{(t)}$ to the server.
\ENDFOR
\STATE \textbf{Server-side}:
\STATE Calculate $C_i^{(t)}$ for each client $i$.
\STATE Aggregate grouped critical model $\boldsymbol{\delta}_i^{(t)}$ for each $i$.
\STATE Aggregate global model $\bar{\btheta}^{(t)}=\frac{1}{N}\sum_{i=1}^N\btheta_i^{(t)}\odot\bm_i^{(t)}$.
\STATE Send combined model $\btheta_i^{(t+1)}$ back to client $i$.
\STATE \textbf{Client-side}:
\FOR{$i=1$ to $N$ in parallel}
\STATE Initialize model with combined parameter set $\btheta_i^{(t)}$.
\ENDFOR
\ENDFOR 
\end{algorithmic}
\end{algorithm}

\begin{table*}[htbp]
\centering
\renewcommand{\arraystretch}{1.3}
\setlength{\tabcolsep}{1.35mm}
\small
\begin{tabular}{@{}cccccccccc@{}}
\hline
\multicolumn{1}{l|}{} & \multicolumn{3}{c|}{Fashion-MNIST} & \multicolumn{3}{c|}{CIFAR-10} & \multicolumn{3}{c}{CIFAR-100} \\ \hline
\multicolumn{1}{c|}{Method}   & $\alpha=0.1$ & $\alpha=0.5$ & \multicolumn{1}{c|}{$\alpha=1.0$} & $\alpha=0.1$ & $\alpha=0.5$ & \multicolumn{1}{c|}{$\alpha=1.0$} & $\alpha=0.01$ & $\alpha=0.1$ & $\alpha=0.5$ \\ \hline
\multicolumn{1}{c|}{Separate} &95.65$\pm$1.19&88.52$\pm$0.34&\multicolumn{1}{c|}{85.30$\pm$2.02}&81.97$\pm$1.33&60.78$\pm$1.04& \multicolumn{1}{c|}{52.42$\pm$2.89} &81.32$\pm$2.26&46.93$\pm$1.70&23.43$\pm$1.17 \\
\multicolumn{1}{c|}{FedAvg}   &\underline{97.05}$\pm$1.21&\underline{92.25}$\pm$0.51&\multicolumn{1}{c|}{89.55$\pm$1.00}&84.15$\pm$1.77&66.33$\pm$1.96& \multicolumn{1}{c|}{61.93$\pm$1.69}&\underline{83.60}$\pm$3.17&51.47$\pm$2.07&33.67$\pm$0.96 \\
\multicolumn{1}{c|}{FedPer}   &96.78$\pm$1.02&92.02$\pm$0.68&\multicolumn{1}{c|}{\underline{89.85}$\pm$1.59}&84.43$\pm$1.69&66.75$\pm$0.83& \multicolumn{1}{c|}{59.92$\pm$1.13} &\textbf{84.02}$\pm$0.70&50.34$\pm$1.76&25.67$\pm$1.19 \\
\multicolumn{1}{c|}{FedBN}    &96.83$\pm$1.15&92.23$\pm$0.38&\multicolumn{1}{c|}{89.77$\pm$0.83}&85.52$\pm$1.66&69.47$\pm$1.08& \multicolumn{1}{c|}{62.62$\pm$2.02} &82.63$\pm$2.62&\textbf{54.98}$\pm$1.88&\underline{34.78}$\pm$0.72 \\
\multicolumn{1}{c|}{pFedSD}         &\textbf{97.17}$\pm$0.94&\textbf{92.80}$\pm$0.79& \multicolumn{1}{c|}{\textbf{90.23}$\pm$1.15} &\textbf{86.48}$\pm$1.48&\underline{71.10}$\pm$0.22& \multicolumn{1}{c|}{\underline{63.23}$\pm$2.46} &82.07$\pm$2.20&52.93$\pm$2.15&31.55$\pm$0.53 \\
\multicolumn{1}{c|}{FedCAC}   &96.80$\pm$1.18&92.12$\pm$0.60&\multicolumn{1}{c|}{89.42$\pm$0.95}&84.75$\pm$1.31&67.38$\pm$1.06& \multicolumn{1}{c|}{61.97$\pm$1.63} &82.87$\pm$3.13&52.47$\pm$2.77&33.73$\pm$1.20 \\ \hline
\multicolumn{1}{c|}{FedPURIN}  &96.55$\pm$1.25&92.15$\pm$0.18&\multicolumn{1}{c|}{89.50$\pm$1.14}&\underline{85.90}$\pm$1.53&\textbf{71.53}$\pm$1.29& \multicolumn{1}{c|}{\textbf{65.50}$\pm$1.71} &83.07$\pm$2.83&\underline{54.23}$\pm$2.45&\textbf{34.92}$\pm$0.61 \\ \hline
\end{tabular}
\caption{Performance of different methods under Dirichlet non-IID settings on Fashion-MNIST, CIFAR-10, and CIFAR-100 datasets. 
Values are reported as mean $\pm$ standard deviation over three data splits. 
Bold and underline fonts denote the highest and second-highest results, respectively.}\label{tab:acc}
\end{table*}

\begin{figure*}
    \centering
    \subfigure[]{
    \label{a=0.1}
    \includegraphics[width=0.32\linewidth]{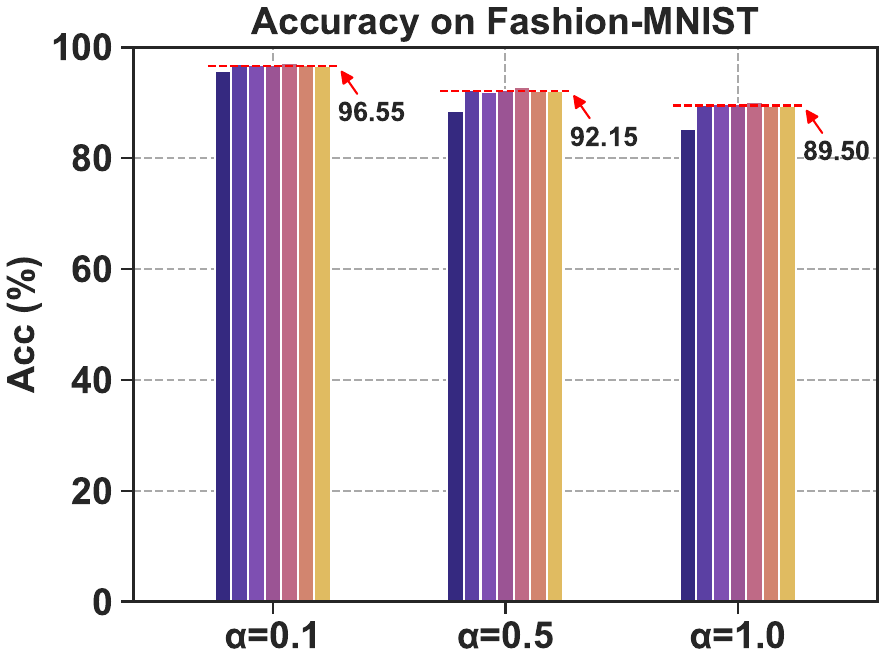}}
    \subfigure[]{
    \label{a=0.5}
    \includegraphics[width=0.32\linewidth]{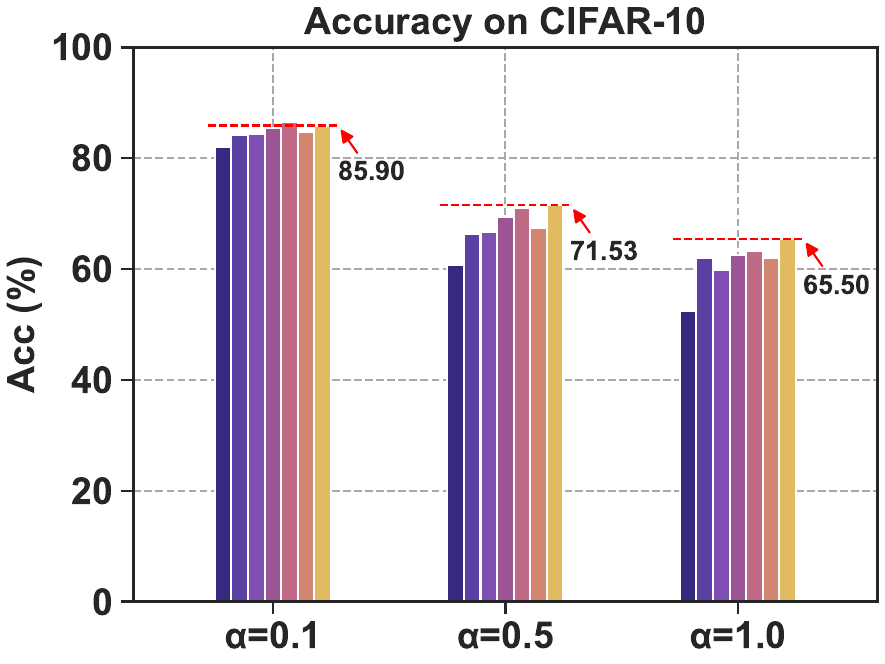}}
    \subfigure[]{
    \label{a=1.0}
    \includegraphics[width=0.32\linewidth]{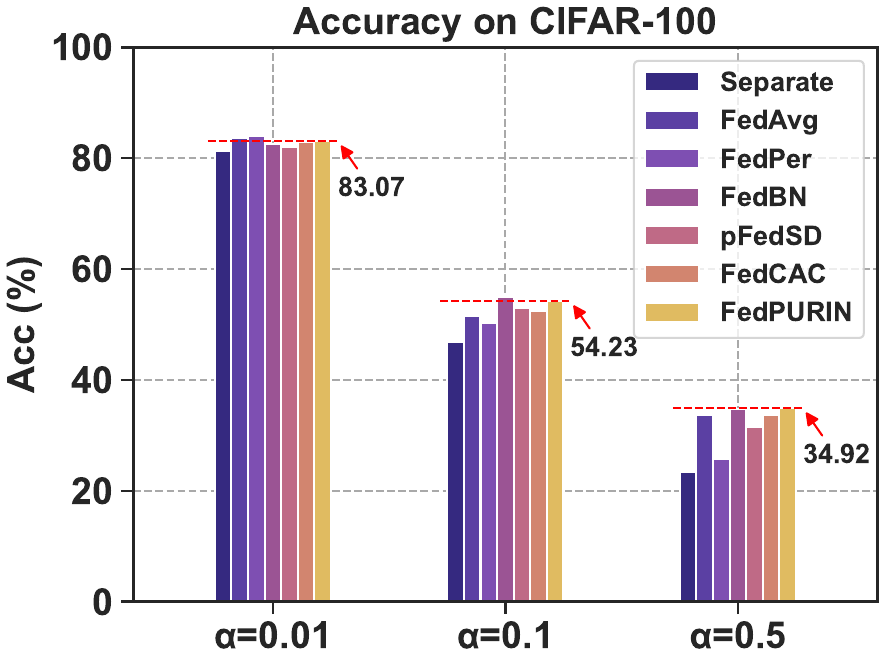}}
    \caption{Comparisons under Dirichlet non-IID on Fashion-MNIST, CIFAR-10, and CIFAR-100 datasets. Dashed lines and arrows annotate the accuracy of FedPURIN.}
    \label{fig:acc}
\end{figure*}

\section{Experiments}\label{sec4}
In this section, we evaluate the proposed FedPURIN framework on standard image classification benchmarks under diverse non-IID settings, focusing on accuracy and communication efficiency. Additional experiments on model components further validate its effectiveness. The results show that FedPURIN achieves competitive performance and substantial communication reduction compared with state-of-the-art methods.
\subsection{Experiment Setup}\label{sec:4.1}
\textbf{Datasets.} 
We evaluate FedPURIN on three datasets: Fashion-MNIST~\cite{xiao2017fashionmnistnovelimagedataset}, CIFAR-10, and CIFAR-100~\cite{krizhevsky2009learning}. To validate the effectiveness of our method under the non-IID scenario, we adopt the \textbf{Dirichlet non-IID} setting, which is commonly used for PFL~\cite{lin2020ensemble, kim2022multi}. Each client's data are drawn from the Dirichlet distribution \emph{Dir($\alpha$)} to create disjoint heterogeneous data. The value of $\alpha$ controls the degree of non-IID partition. As $\alpha$ increases, the degree of class imbalance gradually decreases and the number of classes assigned for each client increases. However, a smaller $\alpha$ results in more skewed datasets and more personalized data split. Therefore, the Dirichlet non-IID is capable of evaluating the performance of algorithms in various complex non-IID scenarios. Experiments take Dirichlet parameters $\alpha=\{0.1, 0.5, 1.0\}$ for Fashion-MNIST and CIFAR-10, and $\{0.01, 0.1, 0.5\}$ for CIFAR-100. To guarantee the significance of client collaboration, we follow the setting of previous methods and assign a fairly small amount of data to each client. Only 500 training samples and 100 test samples are held by one client, both have the same distribution.

\textbf{Baselines.}
We compare FedPURIN with four typical methods from PFL scope. \textbf{FedPer}~\cite{arivazhagan2019} is a pioneering PFL method that performs collaboration except for the classification layer, while \textbf{FedBN}~\cite{li2021fedbn} conducts aggregation except for the BatchNorm (BN) leyers. Different from them, \textbf{pFedSD}~\cite{jin2022personalized} fulfills PFL via knowledge distillation and transfer learning. We also report the important baseline \textbf{FedCAC}~\cite{wu2023bold}, which calculates sensitivity and identifies critical parameters for sparse model updates. In addition, we list the result from \textbf{FedAvg}~\cite{mcmahan2017communication} and the pure independent local training method \textbf{Separate}.

\textbf{Implementation details.} 
Experiments are conducted on NVIDIA RTX 4090 GPUs and an Intel Xeon Silver 4314 CPU using Python 3.8.20. For fair comparison, we build upon the public FedCAC repository, adopting its hyperparameters and well-tested baseline implementations. We employ SGD with a 0.1 learning rate, $N=20$ clients, and $E=5$ local epochs. Given datasets varied in complexity, we set maximum communication rounds $T=200$ for Fashion-MNIST/CIFAR-10 and $300$ for CIFAR-100, with collaboration thresholds $\beta=100$ and $170$ respectively. ResNet-8 is used for Fashion-MNIST/CIFAR-10 and ResNet-10 for CIFAR-100. Reported accuracies represent the best result before global aggregation across the communication rounds. To better examine the capability of each method, we record the averaged results over different data splits corresponding to three random seeds. The statistical parameters (e.g., \emph{running\_mean} and \emph{running\_var}) in the BN layers are excluded for each algorithm as in previous works. In addition, the default FedPURIN is equipped with approximated gradient $\Delta\btheta$, no Hessian, and $\tau=0.5$ unless specified. Collaboration of BN layers is also excluded.

\renewcommand{\dblfloatpagefraction}{.9}
\begin{figure*}[htbp]
    \centering
    \subfigure[$\alpha=0.1$]{
    \label{a=0.1}
    \includegraphics[width=0.32\linewidth]{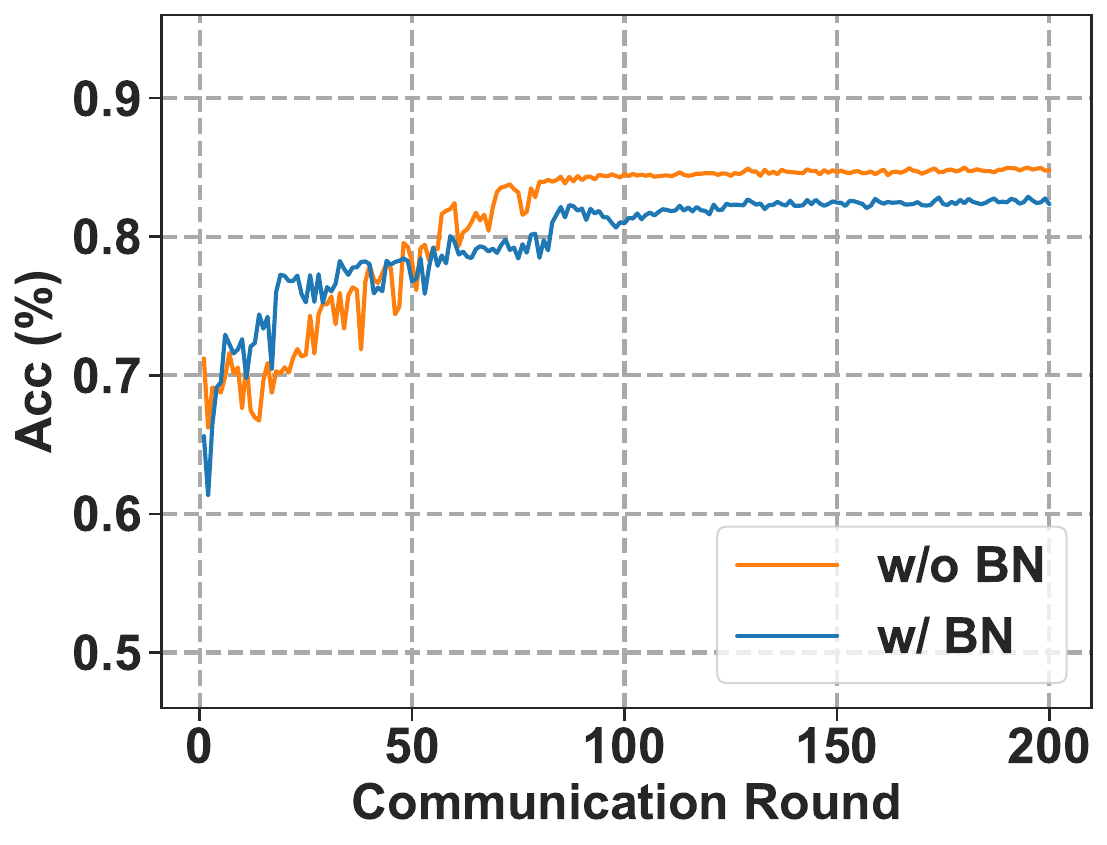}}
    \subfigure[$\alpha=0.5$]{
    \label{a=0.5}
    \includegraphics[width=0.32\linewidth]{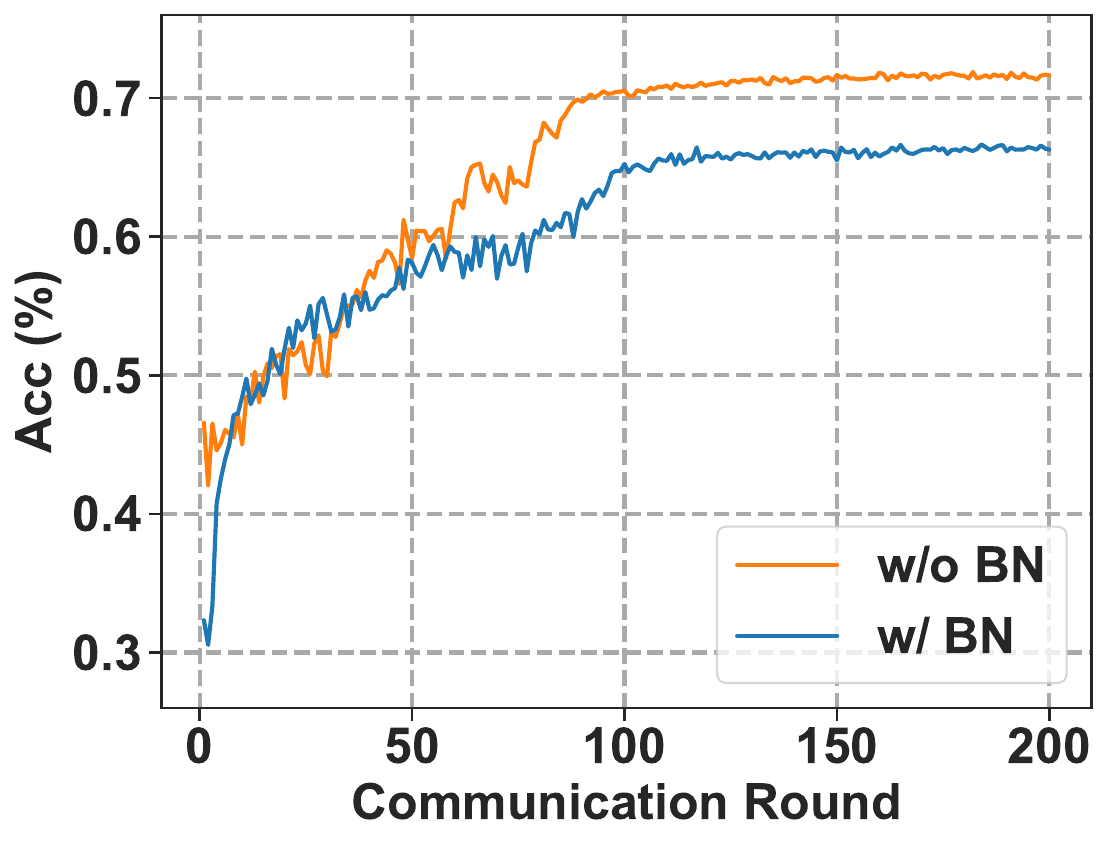}}
    \subfigure[$\alpha=1.0$]{
    \label{a=1.0}
    \includegraphics[width=0.32\linewidth]{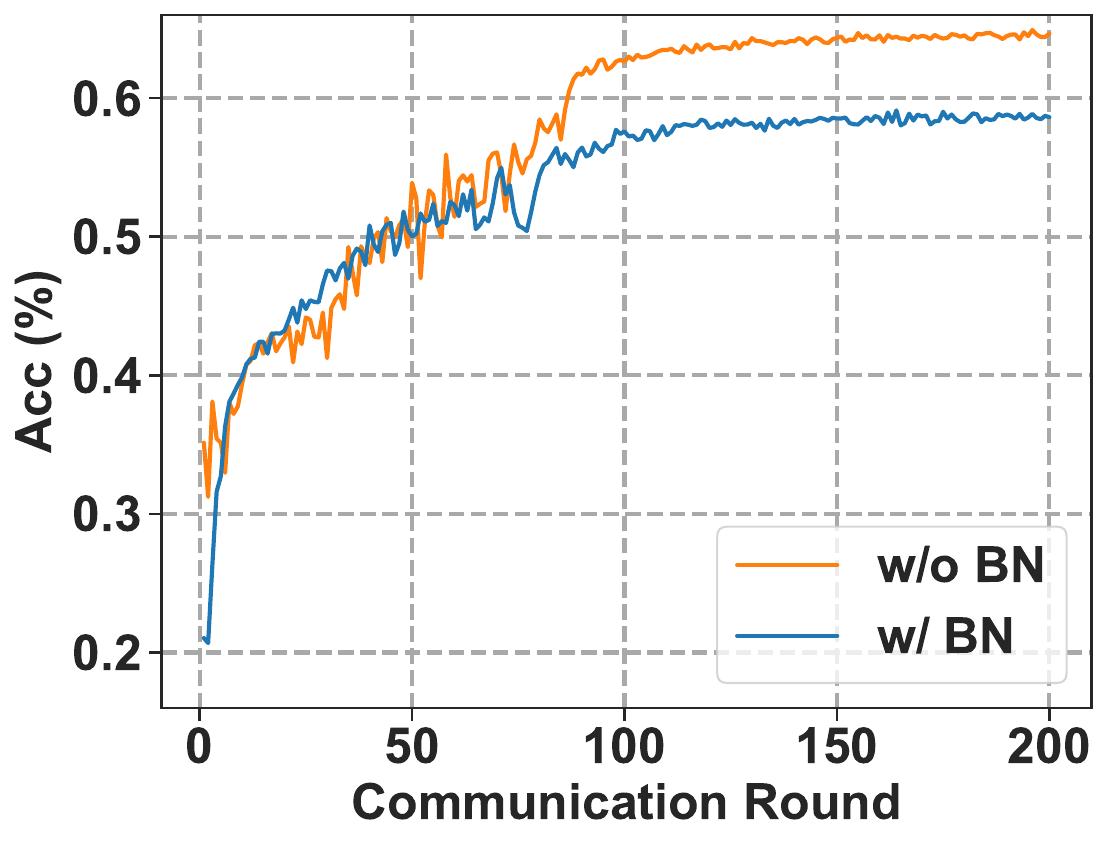}}
    \caption{Accuracy of different BN schemes on CIFAR-10. The solid line represents the average over three runs.}
    \label{fig:bn}
\end{figure*}

\renewcommand{\dblfloatpagefraction}{.9}
\begin{figure*}[htbp]
    \centering
    \subfigure[$\alpha=0.1$]{
    \label{a=0.1}
    \includegraphics[width=0.32\linewidth]{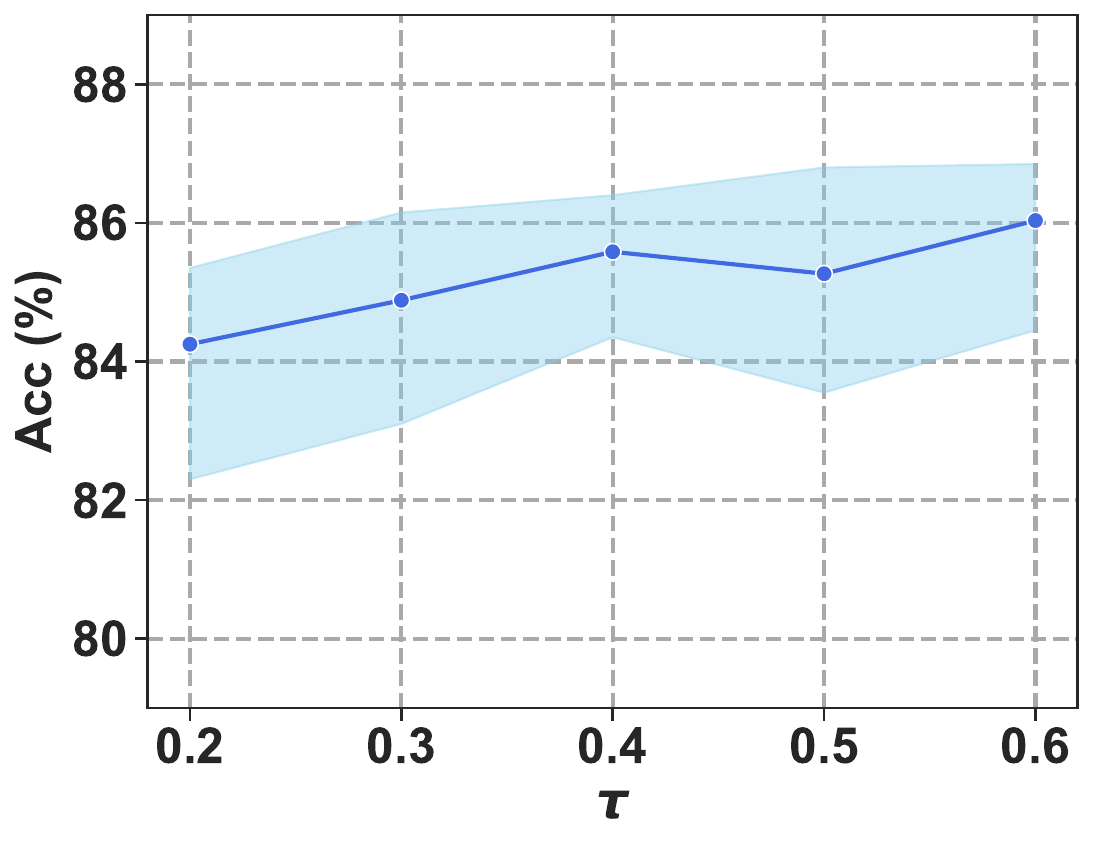}}
    \subfigure[$\alpha=0.5$]{
    \label{a=0.5}
    \includegraphics[width=0.32\linewidth]{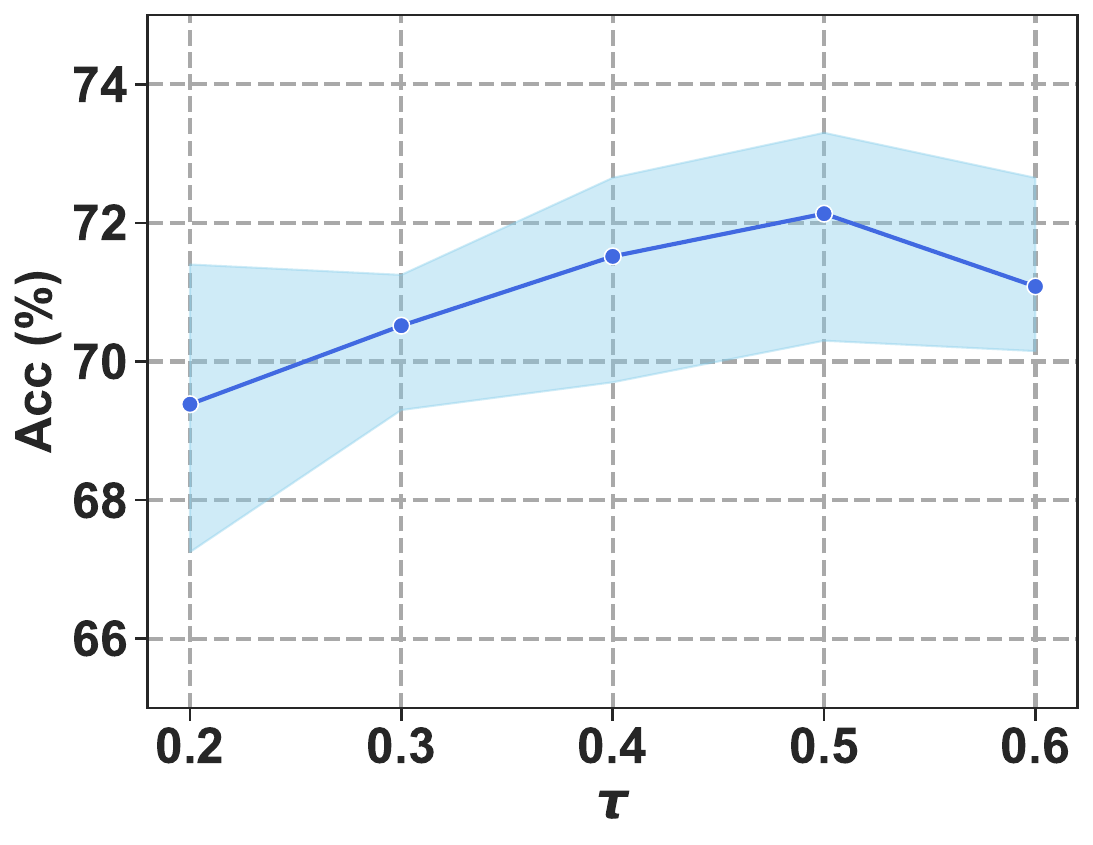}}
    \subfigure[$\alpha=1.0$]{
    \label{a=1.0}
    \includegraphics[width=0.32\linewidth]{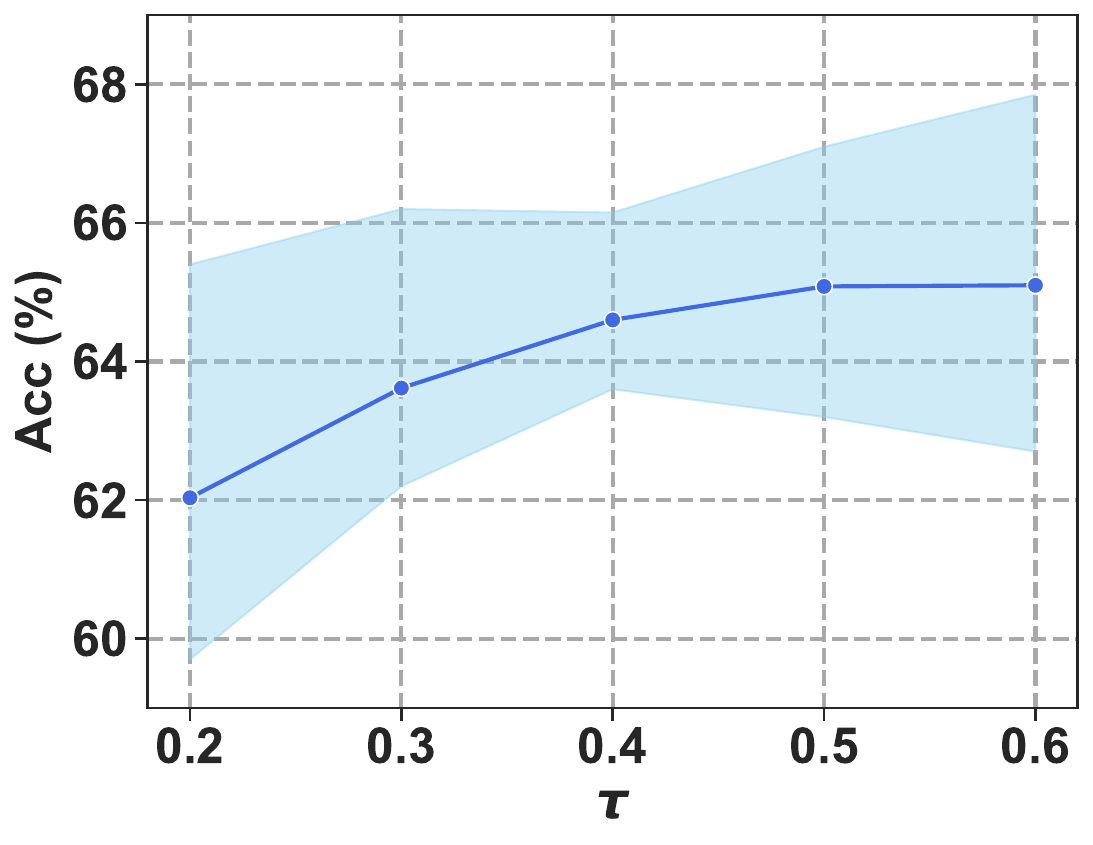}}
    \caption{Accuracy of different $\tau$s on CIFAR-10. The solid line represents the average over three runs, and the shaded area depicts the range of variation (min–max).}
    \label{fig:tau}
\end{figure*}

\subsection{Comparison with SOTA Methods}\label{sec:4.2}
To facilitate a rigorous comparison between FedPURIN and baseline methods, Table~\ref{tab:acc} documents peformance derived from three distinct data splits per dataset, generated through varying random seeds. This approach ensures robustness across heterogeneous data distributions and substantially strengthens the validity of our analysis. We employ the exact gradient calculation in the results presented here, with a comprehensive analysis of this choice provided in the next subsection. The results demonstrate FedPURIN's consistent competitiveness across different scenarios, achieving robust performance that either exceeds or maintains minimal performance gaps against most PFL benchmarks.

Notably, performance characteristics exhibit significant dataset dependence. On Fashion-MNIST, all methods except `Separate' demonstrate comparable accuracy, reflecting limited complexity of the task. For CIFAR-10, pronounced performance disparities emerge between conventional methods and advanced PFL techniques. FedPURIN delivers substantial performance gains at $\alpha=0.5/1.0$, establishing its efficacy in moderately skewed environments. In the more complex CIFAR-100 benchmark, FedPURIN remains competitive with minor deviations from leading alternatives.

To further visualize the closeness among methods, we deploy the barplot in Fig.~\ref{fig:acc}, which shows that all SOTA methods significantly outperform the isolated `Separate' approach, with the performance gap widening under higher $\alpha$ values. FedAvg excels on Fashion-MNIST due to dataset simplicity but lags behind PFL approaches as complexity increases. We highlight the performance of FedPURIN by dashed lines and arrows: it consistently matches or surpasses SOTAs, particularly on CIFAR datasets under exact gradient calculation. Crucially, it exceeds FedCAC (the most relative baseline) at higher $\alpha$ values and harder datasets, confirming the efficacy of our sparse parameter collaboration scheme.

A critical architectural distinction underpins FedPURIN’s success: the deliberate exclusion of BatchNorm layers combined with critical parameter aggregation principles. When directly compared against these conceptually closest relatives, FedPURIN outperforms FedBN in five experimental configurations with only negligible deficits elsewhere, while surpassing FedCAC in eight of nine dataset splits. In short, no method demonstrates absolute dominance across all situations, but FedPURIN always stays in the first-tier part, demonstrating comparable effectiveness to SOTA methods—either exceeding their performance or maintaining negligible gaps. These compelling advantages underscore FedPURIN's dual merits of efficiency and effectiveness, positioning it as a promising solution to real-world distributed learning tasks.

\subsection{Component Ablation Studies}\label{sec:4.3}
In this section, we examine the contribution of different functional modules or hyperparameters in FedPURIN.

\textbf{Independent BatchNorm.}
BatchNorm (BN) layers capture client-specific statistical properties of local data~\cite{li2021fedbn}. The aggregation on such layers should be tested carefully. We conducted experiments on CIFAR-10 to investigate the impact of retaining BN layers locally in federated learning. Here, `w/o BN' denotes FedPURIN without aggregating BN learnable parameters, while `w/ BN' allows their collaboration. Fig.~\ref{fig:bn} shows the average testset performance under three random seeds and different non-IID situations. A consistent advantage for the `w/o BN' scheme exists across non-IID scenarios. The performance gap even widens as the data distributions become more uniform (larger $\alpha$). Aggregating BN layers homogenizes feature representations, thereby impairing client-specific discriminative feature learning for other model parameters. Simultaneously, the increased complexity of multi-class discrimination under larger $\alpha$ renders the model more vulnerable to subtle parameter disturbances, which readily degrades accuracy. Consequently, collaboration restricted to non-BN parameters yields greater benefits than full-model aggregation, particularly under low skewness conditions.

\textbf{Varying sparsity.}
The hyperparameter $\tau$ crucially balances sensitivity to model perturbations and efficiency: overly small $\tau$s risk excluding critical parameters, while large values increase communication overhead. We evaluated $\tau \in \{0.2, 0.3, 0.4, 0.5, 0.6\}$ across three non-IID scenarios. The results in Fig.~\ref{fig:tau} are consistent with our intuition: lower $\tau$ impedes collaboration through sparse parameter selection, simultaneously hindering effective global model aggregation. Accuracy systematically improves with increasing $\tau$s, but an excessively high threshold that incorporates too many elements may lead to a performance decline or higher communication overhead. For a stable and fair comparison, we adopted FedCAC's default parameter $\tau=0.5$ finally, and fixed it for all datasets.

\begin{table}[t]
\centering
\renewcommand{\arraystretch}{1.3}
\setlength{\tabcolsep}{0.66mm}
\small
\begin{tabular}{cc|ccc|ccc|ccc}
\hline
  &   & \multicolumn{3}{c|}{Fashion-MNIST} & \multicolumn{3}{c|}{CIFAR-10} & \multicolumn{3}{c}{CIFAR-100} \\ \hline
$\bg$ & $H$ & 0.1 & 0.5 & 1.0 & 0.1 & 0.5 & 1.0 & 0.01 & 0.1 & 0.5 \\ \hline
  &   & 96.50 & 92.08 & 89.27 & 85.27 & \textbf{72.13} & 65.08 & 82.33 & 53.48 & 33.05 \\
 \checkmark &   & \underline{96.55} & 92.15 & 89.50 & \underline{85.90} & \underline{71.53} & \underline{65.50} & \textbf{83.07} & \textbf{54.23} & \textbf{34.92} \\
  & \checkmark  & 96.48 & \textbf{92.30} & \underline{89.60} & 85.62 & 70.77 & 65.27 & 82.25 & 53.42 & 32.58 \\
 \checkmark & \checkmark  & \textbf{96.57} & \underline{92.27} & \textbf{89.62} & \textbf{86.27} & 71.45 & \textbf{65.63} & \underline{82.55} & \underline{54.03} & \underline{34.87} \\ \hline
\end{tabular}
\caption{Accuracy under different gradient and Hessian terms.}\label{tab1}
\end{table}

\textbf{Different terms of perturbation.} 
FedPURIN employs QIP to compute parameter perturbation for zeroization, leveraging gradient vectors ($\bg$) and Hessian matrices ($H$). Precise $\bg$ computation traditionally requires additional forward passes, incurring computational overhead. Since existing works approximate $\bg$ through parameter variations ($\Delta\btheta$) during local training, we conducted the comparison between it and the exact gradient of the final training batch, which eliminates extra computation without sacrificing accuracy. We further investigated the effect of Hessian term approximated via Fisher information (Eq.~\eqref{eq6}). Controlled experiments across three datasets compared four configurations with/without exact gradients and Hessian terms. Table~\ref{tab1} reports maximum test accuracy averaged over three seeds, revealing comparable overall performance but notable degradation in gradient-approximated methods.

Our systematic comparison yields two main insights. First, the exact gradients are empirically more effective than the stacked variation $\Delta\btheta$ in capturing the loss landscape for parameter selection. With 500 samples per client and 100-sample size of one batch, the final-batch gradients sufficiently depict perturbation dynamics. Second, we observe that the inclusion of the Hessian term does not yield a substantial or consistent improvement in performance. Given that our sparsification ($\tau=0.5$) already filters out a significant portion of the parameters, the marginal benefit of Hessian is outweighed by its computational cost (though little yet in our experiment). Therefore, we conclusively recommend the configuration using exact gradients without Hessian as the preferred choice for practical implementations, affirming the effectiveness and efficiency of our core framework.

\renewcommand{\dblfloatpagefraction}{.9}
\begin{figure*}[htbp]
    \centering
    \subfigure[]{
    \label{layer1}
    \includegraphics[width=0.19\linewidth]{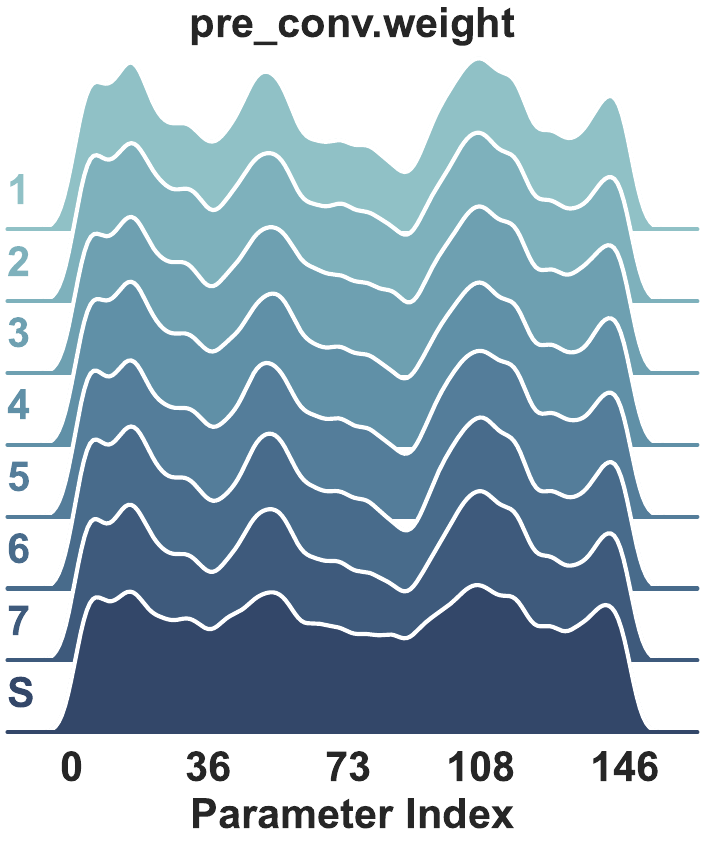}}
    \subfigure[]{
    \label{layer2}
    \includegraphics[width=0.19\linewidth]{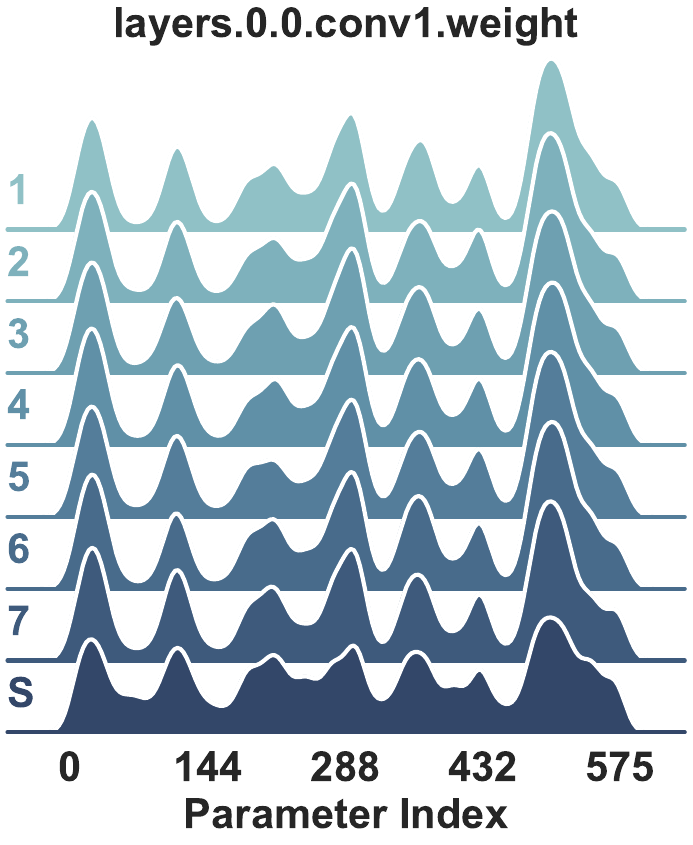}}
    \subfigure[]{
    \label{layer3}
    \includegraphics[width=0.19\linewidth]{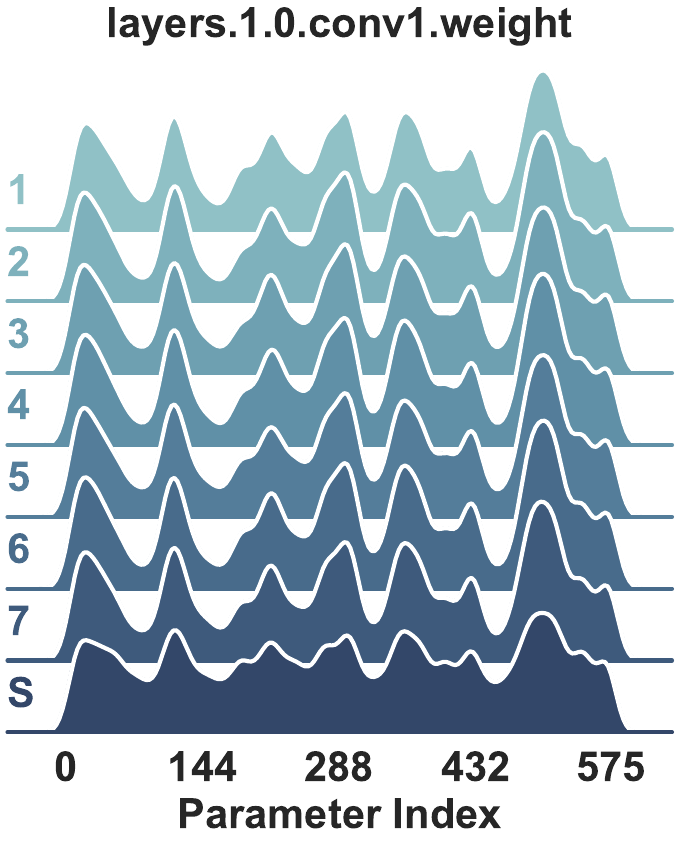}}
    \subfigure[]{
    \label{layer4}
    \includegraphics[width=0.19\linewidth]{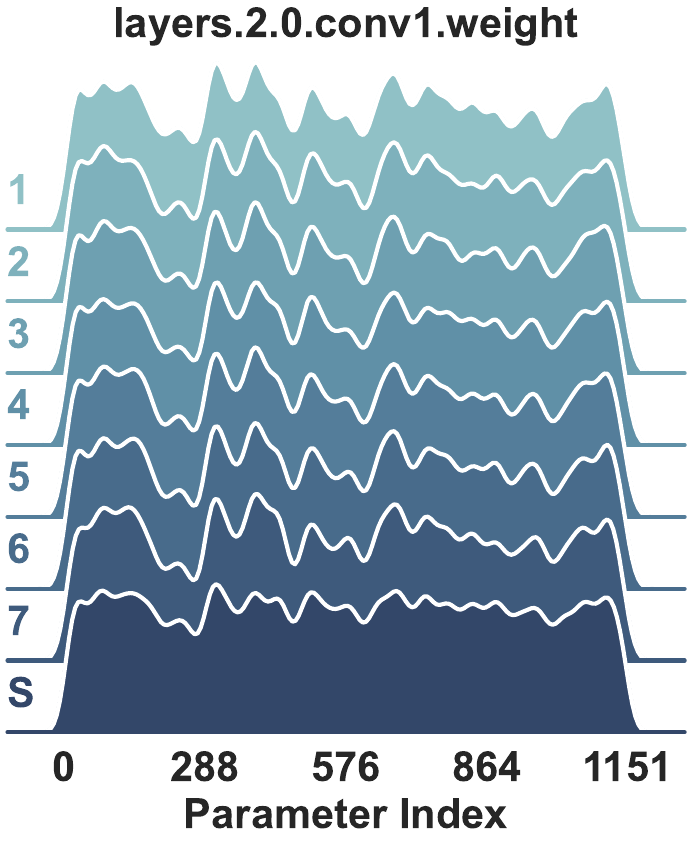}}
    \subfigure[]{
    \label{layer5}
    \includegraphics[width=0.19\linewidth]{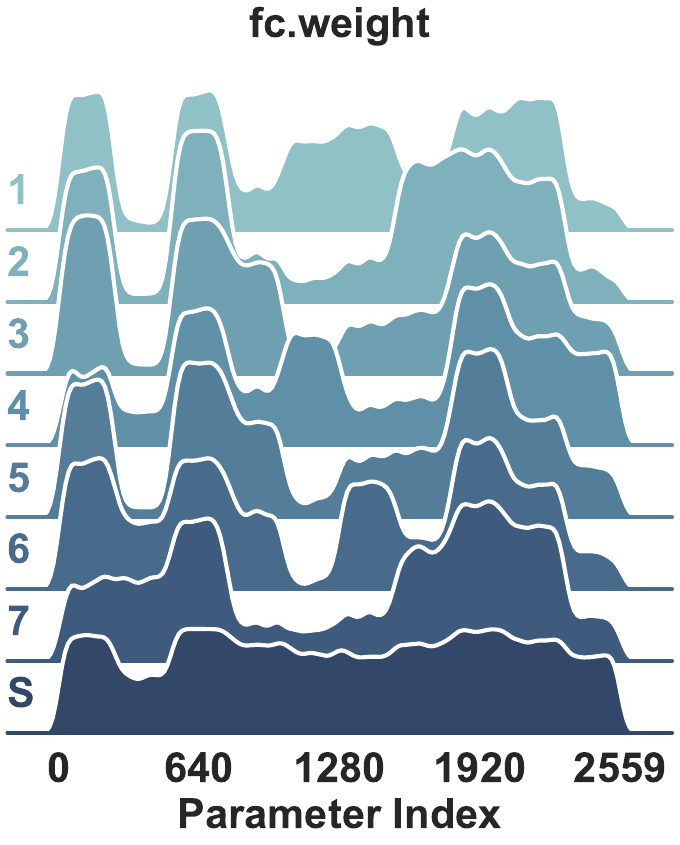}}
    \caption{Distribution of parameter activation frequencies across layers for client and global models during training. Plots of clients (numerically labeled) indicate the fraction of FL rounds each parameter was selected as critical, while the server plot (labeled “S”) reports aggregated non-zero counts.
}
    \label{fig:sparsity}
\end{figure*}

\renewcommand{\dblfloatpagefraction}{.9}
\begin{figure*}[!h]
    \centering
    \subfigure[]{
    \label{heatmap1}
    \includegraphics[width=0.19\linewidth]{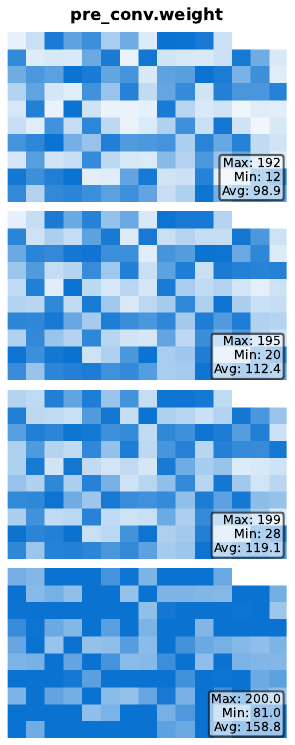}}
    \subfigure[]{
    \label{heatmap2}
    \includegraphics[width=0.19\linewidth]{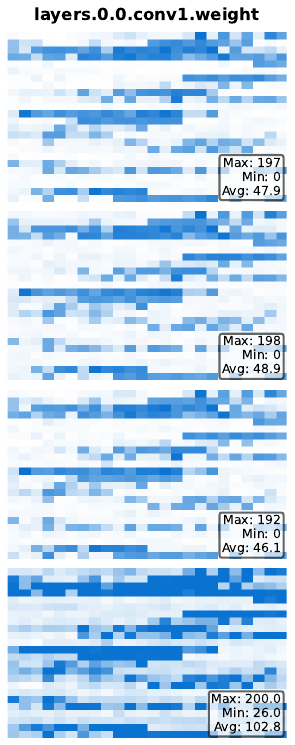}}
    \subfigure[]{
    \label{heatmap3}
    \includegraphics[width=0.19\linewidth]{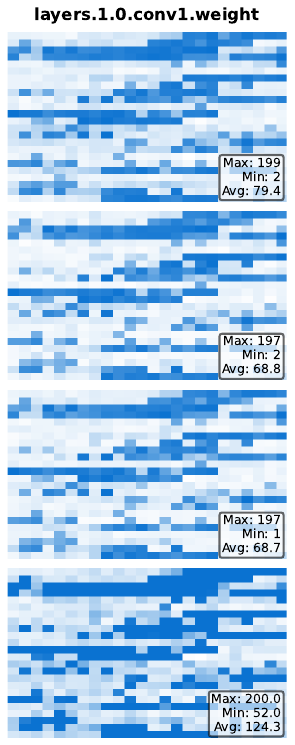}}
    \subfigure[]{
    \label{heatmap4}
    \includegraphics[width=0.19\linewidth]{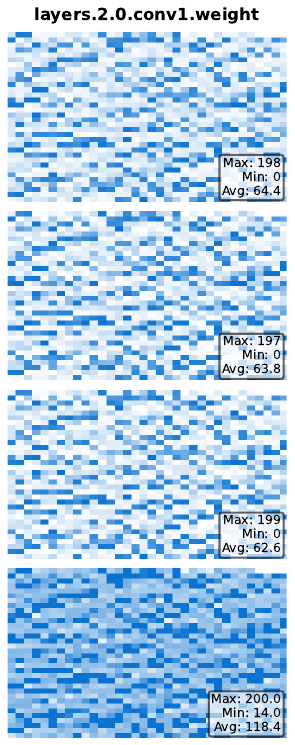}}
    \subfigure[]{
    \label{heatmap5}
    \includegraphics[width=0.19\linewidth]{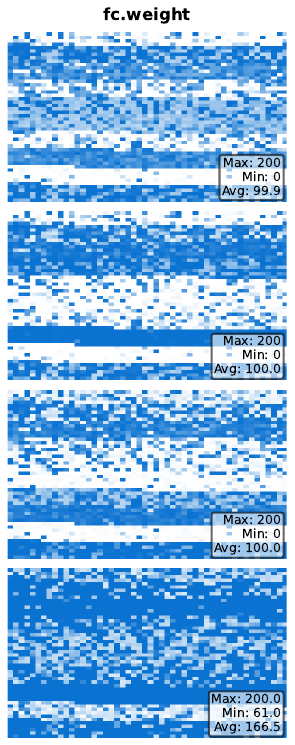}}
    \caption{Heatmaps of parameter activation counts during FL training across layers for client and global models. Rows 1–3 show counts of parameters selected as critical for three clients, while the last row (labeled “S”) shows aggregated non-zero counts for the server model.
}
    \label{fig:heatmap}
\end{figure*}

\subsection{Sparsification Coverage Analysis}\label{sec:4.4}
To characterize dynamic sparsity patterns emerged during federated training, we visualize parameter activation frequencies across client and global models in Fig.~\ref{fig:sparsity}. Experiments are based on the ResNet-8 model trained on CIFAR-10, focusing on the preprocessing convolution (pre\_conv), the first convolutional layer in three BasicBlocks (layers.x.0.conv1), and the classifier (fc). For client models, values indicate the fraction of FL rounds each parameter was selected as critical and uploaded. The server distribution is derived from the non-zero counts in the aggregated global model over the same period. For clarity, we visualize the first output kernel of convolutional layers (e.g., a slice of shape $(1, 3, 7, 7)$ from the full tensor $(Output, Input, Width, Height)$) and the full parameter set of the classifier. The discussion in the following is based on these specific parameters.

Fig.~\ref{fig:sparsity} reveals distinct layer-wise behaviors. For instance, the pre\_conv layer (Fig.~\ref{layer1}) and the final BasicBlock layer (Fig.~\ref{layer4}) exhibit near-uniform parameter selection patterns, suggesting that parameters inside these components are of universal importance. This homogeneity contrasts with the specialized parameter preferences observed in intermediate convolutional layers (Figs.~\ref{layer2}--\ref{layer3}), where collaboration focuses on specific critical parameters, resulting in pronounced distribution peaks. In general, convolutional layers show consistent patterns across clients, reflecting consensus on critical feature extractor parameters. In contrast, the classifier layer (Fig.~\ref{layer5}) shows strongly divergent client-specific patterns, aligned with the non-IID setting and underscoring the need for personalized decision boundaries. This heterogeneity also explains the origins and justification of classifier decoupling strategies such as FedPer~\cite{arivazhagan2019}.

These trends are further elaborated by the heatmaps in Fig.~\ref{fig:heatmap} across model layers. The first three rows correspond to three clients, showing how often each parameter was selected as critical, while the last row represents the non-zero counts of the trivially aggregated server model $\bar{\btheta}$. Darker colors indicate higher counts. Analysis reveals unique functional patterns for clients. The pre\_conv layer (Fig.~\ref{heatmap1}) and the last BasicBlock (Fig.~\ref{heatmap4}) display frequent and relatively uniform parameter activation. By comparison, intermediate BasicBlock layers (Figs.~\ref{heatmap2}--\ref{heatmap3}) exhibit sparser activation with structured patterns, validating the effectiveness of our sparse collaboration scheme. Consensus across clients is evident in the feature extractor layers but breaks down in the classifier (Fig.~\ref{heatmap5}), confirming the heterogeneity induced by non-IID data for the classifier.

As for the server model, it maintains relatively balanced distributions in the ridge plots and exhibits denser patterns in the heatmaps after aggregation, effectively consolidating knowledge from all clients. This ensures the revival of locally zeroized non-critical parameters through the next initialization from the global model. Collectively, these visualizations validate the efficacy of FedPURIN's sparse update mechanism. Clients can focus on critical parameters for personalization, while the server builds an informative global model to support the cooperation of non-critical parameters. As a result, FedPURIN not only improves communication efficiency but also preserves competitive performance, confirming the practical utility of the sparsification mechanism in the field of PFL.

\renewcommand{\dblfloatpagefraction}{.9}
\begin{table*}[htbp]
\centering
\begin{tabular}{c|cc|cc|cc}
\hline
         & \multicolumn{2}{c|}{Fashion-MNIST} & \multicolumn{2}{c|}{CIFAR-10} & \multicolumn{2}{c}{CIFAR-100} \\ \hline
         & $\uparrow$ & $\downarrow$ & $\uparrow$ & $\downarrow$ & $\uparrow$ & $\downarrow$ \\ \hline
FedAvg   & 4.69 & 4.69 & 4.71 & 4.71 & 18.91 & 18.91 \\
FedCAC   & 4.84 & 4.69/2.34 & 4.86 & 4.71/2.36 & 19.06 & 18.91/13.21 \\ \hline
FedPURIN & \textbf{2.38/1.76} & \textbf{4.34/0.69} & \textbf{2.44/1.94} & \textbf{4.31/0.48} & \textbf{6.28/6.19} & \textbf{7.97/1.53} \\ \hline
\end{tabular}
\caption{Average per-round communication overhead (MB) of one client under non-IID setting $\alpha=0.1$. 
Arrows ($\uparrow$, $\downarrow$) denote uplink and downlink volumes. ``a/b'' denotes pre/post-$\beta$ volumes for the critical parameter aggregation scheme.}\label{tab:comm}
\end{table*}

\subsection{Communication Overhead Analysis}\label{sec:4.5}
The sparse critical parameter upload strategy of FedPURIN achieves significant communication efficiency. The overhead statistics were collected from the same run in the main experiment, with only $\bg$ activated. To obtain the precise communication cost for other baseline methods, we performed an independent profiling run under identical experimental configurations and random seeds. Table~\ref{tab:comm} reports the average client communication volumes across random seeds and training rounds on non-IID setting $\alpha=0.1$, counting only parameter transfers (excluding packaging overhead). Only parameter transfers are counted, excluding packaging overhead. Baselines exhibit near-identical overheads as all exchange full models, thus only FedAvg is reported. For FedCAC, critical parameter collaboration halts after $\beta$-th aggregation, reducing downlink to non-critical updates. FedPURIN even cancels the upload of noncritical parameters, further reducing the uplink overhead. In addition, the combined model of critical and global parameters is also sparser than the complete model, which reduces the downlink overhead as well. Here we already considered the extra mask information for FedCAC and FedPURIN by counting 1 bit for each element.

Notably, FedPURIN's uplink overhead reduction deviates from the theoretical ratio $1-\tau$ due to our cutoff operation. When top-$\tau$ perturbation falls below 1e-10, we omitted these values to avoid overly sensitive results. In Fashion-MNIST and CIFAR-10, the algorithm observed a few extreme small non-zero perturbations and removed them. The proposed scheme also effectively filters out the mass of vanishing or zero perturbations produced by the excessively large model trained on CIFAR-100. This filtering directly accounts for the substantially lower transmission volume compared to the baseline methods. Furthermore, it is observed that after the $\beta$-th round, the non-zero parameter positions in the aggregated global model exhibit increased similarity to the clients' critical parameters, which also decreases the downlink transmission. This growing alignment coincides with the general FL training dynamics, where the model stabilizes and the necessary parameter updates across different clients naturally converge toward a consensus.

In summary, FedPURIN reduced the uplink overhead by at least 53.3\% and the downlink overhead by at least 46.3\% for ResNet-8 on two simple datasets and achieved reductions of up to 67.0\% and 72.6\% for ResNet-10 on CIFAR-100. This significant overhead reduction ensures the feasibility of FedPURIN in resource-constrained FL applications.

\section{Conclusion}
In this paper, we propose FedPURIN, a comprehensive framework for personalized federated learning that addresses the critical challenge of balancing model personalization with communication efficiency. Our method decouples client models through a novel integer programming approach that enables theoretically-grounded identification of critical parameters via perturbation analysis. Building on this foundation, we design a communication-efficient aggregation scheme that uploads only these critical parameters for collaboration. Extensive experiments on model accuracy, ablation studies, mask distribution, and communication overhead demonstrate FedPURIN's efficacy in accurately and efficiently accomplishing federated learning tasks across diverse non-IID scenarios. The framework maintains robust performance under high heterogeneity while substantially minimizing communication costs, achieving reductions of 46\% to 73\% compared to existing approaches. This work establishes that sparse, selective parameter collaboration can effectively replace full-model updates without compromising performance, providing both immediate practical benefits for edge intelligence systems and valuable insights for future federated learning research.

\section*{CRediT authorship contribution statement}
\textbf{Lunchen Xie}: Methodology, Writing – original draft, Software, Investigation, Formal analysis, Project administration; \textbf{Zehua He}: Investigation, Writing – review and editing; \textbf{Qingjiang Shi}: Conceptualization, Methodology, Validation, Resources, Supervision, Funding acquisition.

\section*{Acknowledgement}
This work was supported in part by the Science and Technology Commission of Shanghai Municipality under Grant 24DP1500704, and in part by the Fundamental Research Funds for the Central Universities under Grant 22120230311.

\bibliographystyle{elsarticle-num}
\bibliography{bibography}




\end{document}